# ANL: Anti-Noise Learning for Cross-Domain Person Re-Identification

Hongliang Zhang*, Shoudong Han*†, Xiaofeng Pan, Jun Zhao

*Abstract*—Due to the lack of labels and the domain diversities, it is a challenge to study person re-identification in the cross-domain setting. An admirable method is to optimize the target model by assigning pseudo-labels for unlabeled samples through clustering. Usually, attributed to the domain gaps, the pre-trained source domain model cannot extract appropriate target domain features, which will dramatically affect the clustering performance and the accuracy of pseudo-labels. Extensive label noise will lead to sub-optimal solutions doubtlessly. To solve these problems, we propose an Anti-Noise Learning (ANL) approach, which contains two modules. The Feature Distribution Alignment (FDA) module is designed to gather the id-related samples and disperse id-unrelated samples, through the camera-wise contrastive learning and adversarial adaptation. Creating a friendly cross-feature foundation for clustering that is to reduce clustering noise. Besides, the Reliable Sample Selection (RSS) module utilizes an Auxiliary Model to correct noisy labels and select reliable samples for the Main Model. In order to effectively utilize the outlier information generated by the clustering algorithm and RSS module, we train these samples at the instance-level. The experiments demonstrate that our proposed ANL framework can effectively reduce the domain conflicts and alleviate the influence of noisy samples, as well as superior performance compared with the state-of-the-art methods.

## I. INTRODUCTION

Person re-identification (Re-ID) intends to match images of the same identity captured by different cameras. It is applied to the fields such as surveillance and security [1]. With the rapid evolution of deep convolutional neural networks and the construction of large-scale datasets, supervised Re-ID has achieved remarkable results [2]-[5]. Due to the domain differences, the Re-ID models trained on the source domain have a significant performance degradation on the target domain. Obtaining sufficient unlabeled data is effortless, but the tedious data annotation will limit the application of Re-ID in new scenarios. Therefore, this work concentrates on the cross-domain setting, which learns the discriminative feature expression of the target domain by using the labeled source domain data and the unlabeled target domain data. The current methods can be divided into bridging domain gaps and learning target domain features.

The former can continue to be subdivided into feature-level and image-level. For the feature-level, the main idea is the domain feature alignment. MMFA [6] aligned the distribution of mid-level semantic attributes by Maximum Mean Discrepancy. CCE [7] confused the pedestrian features through domain adversarial learning. In contrast, the image-level approaches mainly adopt style transform from the source domain to the target domain while preserving the original identities [8], [9]. However, it is not enough just to reduce the domain differences, neither for feature-level nor image-level. Because the intrinsic properties of the target domain are ignored. For example, the Market1501 [10] was collected in summer while DukeMTMC [11], [12] in winter. The discriminative representation learned from short-sleeved dresses may fail in distinguishing wadded jackets.

The latter methods mainly assign pseudo-labels to unlabeled data by clustering algorithm [13]-[15]. Iterative label assignment and feature training improve the recognition accuracy of the model. In this process, the knowledge of the target domain is considered and the constraints on the samples are more explicit. However, there are three main limitations to existing works. (1) Due to the cross-domain conflicts, it is not reliable to directly utilize the pre-trained source domain model to calculate the sample feature distance in the target domain. (2) Noisy samples produced by inaccurate clustering will significantly impact the final Re-ID performance. Noise reduction is an urgent problem to be considered. (3) The outlier samples generated by the clustering algorithm are likely to be difficult samples that may contain a wealth of information. It is inappropriate to discard them directly, as existing methods do.

We find that the two methods above are complementary to a certain extent. In this paper, we combine the advantages of the two methods and propose a framework called Anti-Noise Learning (ANL) which contains two newly proposed modules, the Feature Distribution Alignment (FDA) module and Reliable Sample Selection (RSS) module. Both can reduce the impact of noise from different perspectives.

This work was supported by the National Natural Science Foundation of China under Grant No. 61105006; Open Fund of Key Laboratory of Image Processing and Intelligent Control (Huazhong University of Science and Technology), Ministry of Education under Grant No. IPIC2019-01; and the China Scholarship Council under Grant No. 201906165066.

* The authors contribute equally to this work. † Corresponding author. H. Zhang, S. Han, X. Pan are the National Key Laboratory of Science and Technology on Multispectral Information Processing, School of Artificial Intelligence and Automation, Huazhong University of Science and Technology, 1037 Luoyu Road, Wuhan, China; P.C: 430074 (e-mail: {hongliangz, shoudonghan, xiaofengpan}@hust.edu.cn).

J. Zhao is with the School of Computer Science and Engineering, Nanyang Technological University, 50 Nanyang Avenue, Singapore 639798 (e-mail: junzhao@ntu.edu.sg).

A well-trained model can extract the discriminative source domain features. But when applied to the target domain, the ability will degrade. The extracted features are inaccurate and the sample distribution is messy, which will lead to sub-optimal clustering results. Reducing domain differences and aligning feature distribution can solve the problem. Combining the advantages of bridging domain gaps methods, our FDA module adjusts the cross-feature space of the target domain samples using camera-wise contrastive learning and adversarial adaptation. After alignment, the id-related samples will be more concentrated and id-unrelated samples will be more scattered.

Keeping the network structure, initialized parameters, and loss functions unchanged, the performance of pseudo-labels is much worse than the real labels (with supervision). PENCIL [16] train network parameters and probability labels in a framework. It corrected the noisy labels by networks. But it may incorrectly correct the original correct label. Inspired by the method, we proposed a Reliable Sample Selection module. An Auxiliary Model is required to start from a small number of clean samples and gradually grasp the entire data distribution. Through the training and correcting of noisy labels, reliable samples are selected and provided to the Main Model. It solves the noise accumulation problem of conventional reliable sample selection based on distance. In this process, the unreliable samples filtered out will be regarded as outliers. At the same time, the clustering algorithm will also produce outlier samples. For the outlier samples, different from the existing methods that directly discarded them [17]-[19] or simply assigned nearest neighbor labels [20], we adopt the way of the instance to let them participate in training and extract their information.

In summary, the contributions of this work are three-fold:
- We introduce the Feature Distribution Alignment module, which directly optimizes the target domain feature distribution. It can reduce the distance of the id-related samples and create a friendly condition for subsequent clustering.
- We propose a Reliable Sample Selection module. An Auxiliary Model corrects noisy labels and provide reliable samples to the Main Model. The problem of noise accumulation is alleviated.
- We utilize outlier samples in the instance way. The diversity of training samples is guaranteed while the noise is reduced.

## II. RELATED WORK

### A. Unsupervised Person Re-identification

Although supervised Re-ID have achieved impressive results [21]-[24], the lack of labels still limits the development and application of Re-ID. Traditional unsupervised Re-ID extract manual features [25]-[27] to build the transition matrix. However, the manual features are often not robust while the matrix is difficult to train. Recently, unsupervised Re-ID based on deep learning make remarkable progress compared with traditional methods. One of the methods is based on reducing the domain difference, which can be divided into two perspectives: image-level and feature-level. As for image-level, DA-2S [28] reduced the gaps by generating images that suppress the background instead of hard image segmentation. CR-GAN [29] followed the principle of "divide-and-conquer" and transferred the image style from the perspectives of illumination, resolution, and camera-view while preserving the original identities. But the quality of the generated images seriously affects the training process. Compared with our methods, they simply learned the apparent information of the generated images and neglected to dig the inherent properties of the target domain. Regarding feature-level, ARN [30] extracted the domain-invariant features through the encoding and decoding network to eliminate the domain differences. Some methods try to use the relationship between the source domain and the target domain as soft-label to guide training [31], [32]. CSCLP [31] constructed soft-labels by mining the semantic similarity between the target images and generated images. But the fuzzy similarity was hard to be strong guidance.

At present, the clustering-based methods [15], [18], [20], [33] perform well, trained by iterative density-based clustering and discriminative feature learning. SSG [18] introduced local features and captured richer semantic information through both global and local information. But existing works directly use the pre-trained source domain model to extract the feature of target instances, while the distribution of the two domains is not aligned. Errors occur at the beginning of clustering. We merge the advantages of bridging domain gaps methods to solve the problem. Specially, we utilize camera-wise contrastive learning and adversarial adaptation to gather similar samples. For the label noise problem, selecting reliable samples is a common solution to noise problems [34]-[37]. DCML [35] determine whether the sample is clean by the distance between it and the cluster center which belongs to. However, like the clustering algorithm, the selection criterion is based on distance. Once the distance measurement fails, the sample selected by this method may not be reliable, that is, the error accumulation problem. On the contrary, our method uses an Auxiliary Model to correct labels and to select reliable samples, thereby avoiding the problem of error accumulation. Besides, in order to avoid the negative impact of directly discard unreliable samples or outliers, we train them in the instance way, by which the diversity of training samples is guaranteed. The effect on several datasets proves the advantages of our method.

### B. Contrastive Learning

The internal structure of data contains more information than the label. We can learn knowledge from the data structure under the condition without labels. A popular paradigm is to construct positive and negative sample pairs separately, then constraint the positive sample pairs closer than the negative. N-pair Loss [38] extended the commonly used triple loss to n-pair negative pairs, achieving efficient convergence meanwhile with low computation cost. DMI [39] maximized the mutual information between input data and the learned high-level representations, achieving the consistency of global and local information. ECN [40] introduced three underlying invariances to learn the intra-domain variations in the target domain. But it only paid attention to the global feature distribution and ignored the feature relationship information of the camera subdomain. Our method uses camera relations and proposes camera-wise contrastive learning. Through the contrastive learning, the feature distribution can be aligned. The id-related samples will be closer and id-unrelated samples will be more dispersed, which is beneficial to create an appropriate clustering condition.

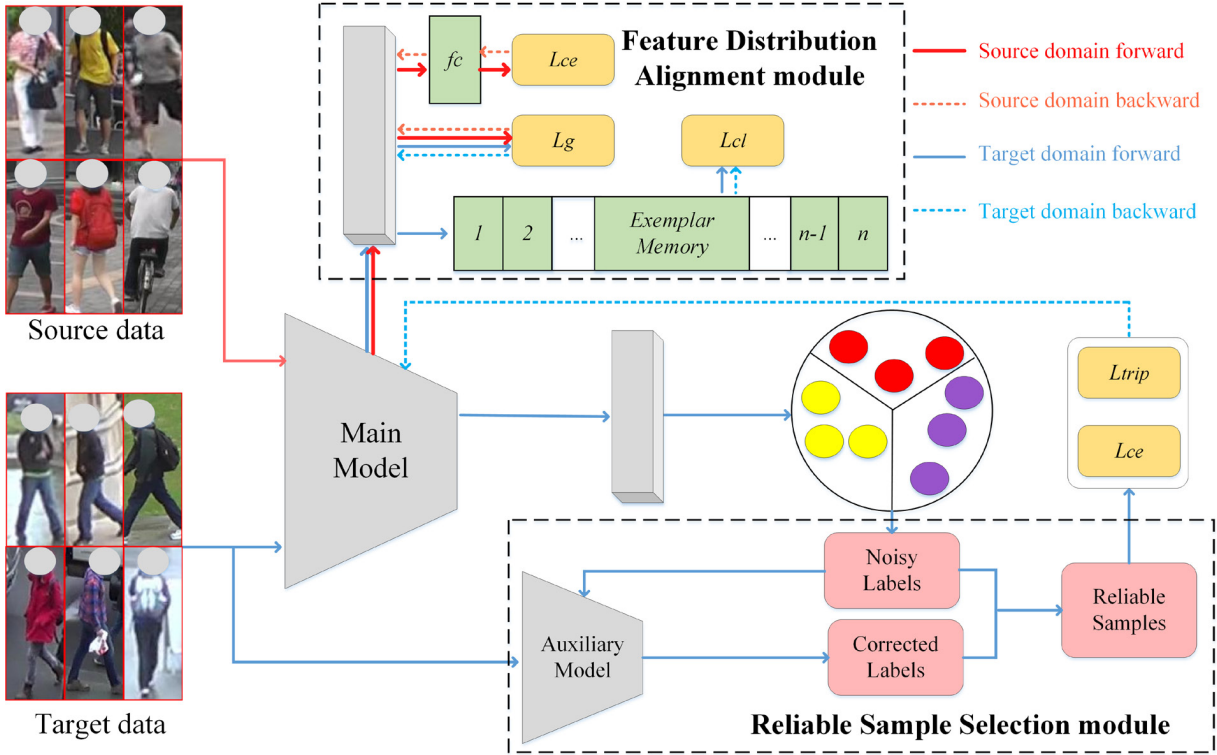

Fig. 1. The framework of the proposed method. In the Feature Distribution Alignment module, the features of the two domains are extracted through a shared parameters backbone network. The source flow uses cross-entropy loss and the target flow is trained by contrastive learning. Besides, an adversarial adaptation loss is used to reduce domain differences. The Main Model is trained by iterative density-based clustering and discriminative feature learning. In the Reliable Sample Selection module, the Auxiliary Model trains noisy labels and finally screens reliable samples. For more detailed content of RSS, please refer to Fig. 2.

### C. Label Noise Learning

Label noise is a common problem in datasets. There are three main methods to solve the problem: Noisy Channel, Dataset Clean, and Robust Losses. For the Noisy Channel, Goldberger et al. [41] used EM [42] to iteratively train the network and estimate the noise transition matrix. But as the number of classes increases, the calculation of the matrix becomes more intractable. The Dataset Clean approaches cleaned labels [43] or choose the clean samples [44], but usually, a clean auxiliary dataset is required. As for Robust Losses, [45] proved that the mean absolute value of error is more noise-tolerant than categorical cross-entropy loss. A loss which combined the robustness of MAE and the implicit weighting of CCE was suggested in [46]. The robust loss requires sophisticated design and is limited in the face of various noises. PENCIL [16] train network parameters and probability labels in a framework. It corrected the wrong labels by networks. But this method is not suitable for the Re-ID, because the pedestrian identity categories in Re-ID is much larger than the label noise learning problem. It is easy to change the original correct label to the wrong one. The training process on the entire noisy datasets will increase the risk of errors. In this paper, we use an iterative training strategy, which starts with a small and clean subset, learns and assigns labels, and increases the number of training samples. Finally, a sample filtering mechanism is adopted to avoid the negative effects of failed labels learning.

## III. THE PROPOSED METHOD

### A. Preliminary

In the field of unsupervised Re-ID, there are source domain data $\{X_S, Y_S\}$, including $N_S$ images and $I_S$ different identities. Each person image $x_i^s$ is associated with an identity $y_i^s$. The target domain data $X_T$ contains $N_T$ images, whose identity annotations are not available. Our goal is to learn a discriminative representation for the target set.

### B. Overview

The framework of our method is shown in Fig. 1. The objective of the FDA module is to align the target domain feature distribution, facilitating the subsequent clustering process. The cross-entropy loss is used for the source domain while the contrastive learning is for the target domain. An adversarial adaptation function is utilized to obfuscate the feature expressions and map the features of different domains to a common space. The Main Model allocates pseudo-labels and makes feature training. The RSS model is shown in more detail in Fig. 2, an Auxiliary Model is trained to correct labels and filter out noisy samples. So that the Main Model is trained under more reliable samples which can lead to a more robust model. Next, we will describe our method in detail.

*C. Feature Distribution Alignment Module*

The current methods train models in the source domain, making the networks able to extract pedestrian features. However, due to the domain differences, the Re-ID models trained on the source domain have a significant performance drop on the target domain, because the model cannot extract appropriate features for the target domain samples, which can cause a lot of errors in estimating pseudo-labels. It is necessary to train the target domain samples at the same time. In this module, the labeled source data and unlabeled target data are fed-forward into the shared parameters network.

*1) Source Data Learning:* For the source data $x^s$, using the labels as a guide, we treat it as a classification problem like many existing methods. The cross-entropy loss is adopted, formulated as

$$L_{ce} = -\sum_{i=1}^{N_S} y_i^s \cdot \log \widehat{y_i^s} \quad (1)$$

where $\widehat{y_i^s}$ is the predicted probability that the source image $x_i^s$ belongs to identity $y_i^s$.

*2) Target Data Learning:* For a classification problem, the highest responses of the classifier are all visually correlated. It is the apparent similarities instead of semantic labels that make some classes closer than others [47]. Based on this theory, we utilize contrastive learning to mine the rich information contained in the data structure. Specifically, we treat each sample as an individual with a different identity. Through image transformation, we can generate variant image $\tilde{x}_k^t$ for each target image $x_k^t$ ($k = 1 \ldots N_T$). By making the original image close to the variant image, and far from other images, we can learn individual discriminative features. But there are two problems. First, we should calculate $2N$ features each time. Besides, only $N$-1 negative samples can be pushed away at a time ($N$ is the batch size). Uneven sampling or unbalanced distribution of data will cause serious interference. Second, some samples belong to the same category, if we treat each sample as a separate class will introduce noise. For the first problem, a storage unit is designed to store and update the features. Each variant feature $\widetilde{f_k^t}$ will be stored in this cell. By taking advantage of the storage unit, all other samples can be pushed away at a time and the computation cost is reduced. For the second problem, features of the same identities will be closer, so we deem that the nearest neighbors belong to the same identity. Because of the illumination, background, and viewpoint variation, the average pairwise similarity of cross-camera matching is smaller than that of intra-camera matching [48]. If we treat the cross-camera and intra-camera as equal, we may only mine the intra-camera samples. So we subdivide the calculation of similarity into cross-camera and intra-camera. Using $C_i(x_k^t, r_1)$ to represent the set of $r_1$ intra-camera samples with the highest similarity to $x_k^t$. The $C_o(x_k^t, r_2)$ is similar but contains the cross-camera samples. The $s_{ij}$ represents the probability that the $j$-th sample belongs to the class $i$, which is calculated by cosine distance

$$s_{ij} = \begin{cases} \frac{f_i^t \widetilde{f_j^t}}{|f_i^t||\widetilde{f_j^t}|} & i \neq j \text{ and } j \in C_i(x_i, r_1) \cup C_o(x_i, r_2) \\ 1 & i = j \\ 0 & else \end{cases} \quad (2)$$

The objective of contrastive learning is

$$L_{cl} = -\frac{1}{N_T} \sum_{i=1}^{N_T} \sum_{j=1}^{N_T} s_{ij} \log \frac{\exp(f_i^t \widetilde{f_j^t}/\tau)}{\sum_{k=1}^{N_T} \exp(f_i^t \widetilde{f_k^t}/\tau)} \quad (3)$$

where $\tau \in (0,1)$ is a temperature factor that balances the scale of distribution. $\tilde{f}_k^t$ is the feature of variant image $\tilde{x}_k^t$, which is stored in the $k$-th unit of the storage. The feature is updated through

$$\tilde{f}_k^t \leftarrow \alpha \tilde{f}_k^t + (1-\alpha) f_k^t \quad (4)$$

Both the source domain and target domain are considered in the above process. Compared with only trained in the source domain, the model can learn the target domain trait. The distance between samples in the target domain can be better measured, which will facilitate subsequent pseudo-label estimation process.

*3) Adversarial Adaptation:* To further narrow the domain gaps, we adopt the domain adversarial function to obfuscate the distribution of features $f = G(x)$. For convenience, we omit the subscript. The backbone network serves as generator $G$ while a simple perceptron network as discriminator $D$. The loss for the generator is

$$L_g = E_{x \sim T}[(D(G(x^t)) - 1)^2] \quad (5)$$

Combine the above-mentioned losses, the final loss for the FDA module is formulated as

$$L_{FDA} = L_{ce} + L_g + L_{cl} \quad (6)$$

The loss for discriminator is

$$L_d = E_{x \sim S}[(D(G(x^s)) - 1)^2] + E_{x \sim T}[(D(G(x^t)))^2] \quad (7)$$

*D. Reliable Sample Selection Module*

We follow a robust and reliable adaptation framework [14] to cluster the samples. The density-based clustering algorithm [49] is adopted to assign samples into different groups. Through the clustering algorithm, we obtain noisy pseudo-labels. After combining the advantage of reducing domain gaps methods, the accuracy of the pseudo-labels has been improved, but the label noise problem still needs to be solved. Inspired by PENCIL, we start from the perspective of label correcting to reduce the noise accumulation problem generated in the original clustering framework. However, there is a risk of correcting the original correct labels. So referring to the results of clustering pseudo-labels, we filter out some unreliable samples, which will be trained in the way of instance. It is worth noting that the total sample selection process is carried out on an Auxiliary Model, which is responsible for providing more reliable samples to the Main Model. The Main Model optimizes network parameters to improve the recognition effect.

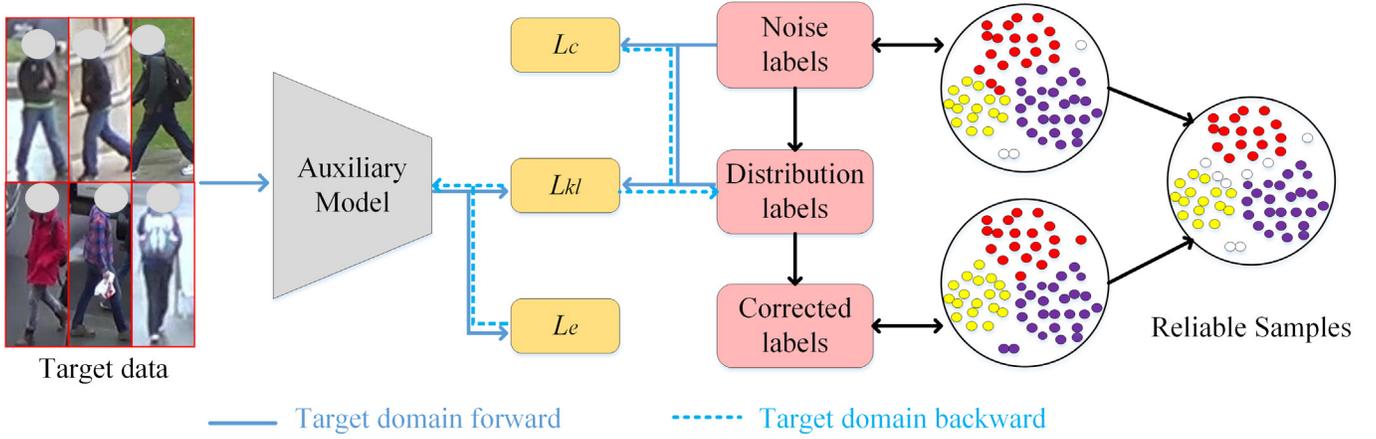

Fig. 2. The framework of the Reliable Sample Selection module. A newly initialized Auxiliary Model starts from a small number of clean samples, grasp the data structure of the target domain. After that, the Auxiliary Model relies on its own discrimination and the noisy labels obtained from the Main Model to learn and correct the labels. The entropy loss instructs the model training while the KL loss corrects the label. In order to reduce the risk of error correction, the corrected samples are regarded as unreliable samples and participate in subsequent training in the form of instances.

Deep models memorize easy instances at first and gradually adapt to hard instances as the training epoch increases [50]. Based on this theory, the entire training process should be like this. At the early stage, the model fits some easy samples based on the labels and obtains the ability to distinguish instances. In the middle period, the model mines the data structure and makes its judgments on the samples, but it is still restricted by the labels. At the end of the training, the difficult samples are learned and the model is completely fitted to the labels. Because of the presence of noisy labels, starting from the middle period, the noisy labels begin to harm the model. In this sub-training process, a two-stage training strategy is proposed to correct noisy labels and select reliable samples.

*1) Auxiliary Model Initialization:* In the first stage, the model is trained with noisy labels, to obtain the ability to distinguish samples. PENCIL [16] implement it on the entire noisy dataset, which will greatly increase the noisy labels ratio. The model is prone to fit the noisy labels and thus cannot be corrected to the correct labels. To avoid it, our method starts with a small number of clean samples, and iteratively selects samples to improve model discrimination. Through the clustering algorithm, we have obtained the non-outlier samples assigned with the pseudo labels $Y^c$. Because the subsequent processes are all applied to the target domain, so we omitted the subscript "$t$" in the following text. Like many existing methods, we select $k$ samples for each pseudo-label class and assume that they are more reliable. The selection principle is the distance from the sample to the corresponding cluster center and the cluster center is the mean of all samples belong to the category. All the selected samples form a labeled set $S_l$ while others form an unlabeled set $S_u$. For the $S_l$ set, cross-entropy and batch-hard triplet loss [51] are used. The entropy loss is adopted for $S_u$ set. In each iteration, the $S_l$ and $S_u$ will change. The output of the network classifier can represent the confidence of the sample discrimination. Samples with a confidence higher than $\lambda$ will be added to $S_l$, and the rest of the samples will be added to $S_u$. As the training progresses, the samples of $S_l$ gradually increase. Through the first stage, the model gains the ability to distinguish samples and the progressive training method reduces the impact of original noise labels.

*2) Labels Correction and Samples Selection:* In the second stage, we train model parameters and noisy labels in a framework. The noisy label $Y^c$ is a hard-label and a one-hot variable. To reduce the noise, we first model the label as a probability distribution and the sum of the probabilities is equal to 1. The hard-label and the probability distribution are presented by $\bar{y}^c$ and $\tilde{y}^c$.

The KL-divergence is often adopted to measure the differences between the two distributions. It is an asymmetric function. Assume that we have two probability distributions, the true distribution $P$ and the estimated distribution $Q$. There are two KL losses

$$KL(P||Q) = \sum P(x)\log\frac{P(x)}{Q(x)} \quad (8)$$

$$KL(Q||P) = \sum Q(x)\log\frac{Q(x)}{P(x)} \quad (9)$$

Equation (8) restricts $Q$ and $P$ to be consistent (both have high probability density) in all areas with high probability density of $P$. But Equation (9) restricts $Q$ and P to be consistent (both have low probability density) in the region where the probability density of $P$ is little. In other words, Equation (8) pays more attention to high probability events while (9) to low probability events.

We are concerned about high probability events because they indicate the category to which the samples belong. As is mentioned above, noisy labels cause side effects in the middle period. To correct the labels, the label $\tilde{y}^c$ is served as the real distribution $P$, and the predicted distribution $z = H(f)$ is regarded as the estimated $Q$. $H$ is a $c$ classifier and $c$ is the number of categories.

The loss function is expressed as

$$L_{kl} = \frac{1}{N_T}\sum_{i=1}^{N_T}\sum_{j=1}^{c} z_{ij}\log(\frac{z_{ij}}{\tilde{y}_{ij}^c}) \quad (10)$$

The $\tilde{y}^c$ is initialized by $\bar{y}^c$ as follow

$$\tilde{y}^c = softmax(\mu \bar{y}^c) \quad (11)$$

where $\mu$ is a large constant, to make $\tilde{y}^c$ closer to a unimodal probability distribution, rather than a uniform distribution.

Because most of the labels are clean, only a few labels are noisy. If our training is unreasonable, we may contaminate the original correct labels. Therefore, we should make the hard-label $\tilde{y}^c$ and the probability distribution $\tilde{y}^c$ similar to a certain extent. The classic cross-entropy loss is suitable for this situation

$$L_c = -\frac{1}{N_T}\sum_{i=1}^{N_T}\sum_{j=1}^{c}\bar{y}_{ij}^c \log \tilde{y}_{ij}^c \quad (12)$$

To train the network to process the noisy labels, the predicted distribution $z$ can be regarded as a guide for $\tilde{y}^c$. Because of the unavailability of labels, the training of $z$ lacks guidance. The entropy loss can force the distribution to peak at only one category rather than be flat at many classes. We want the samples to be classified in a unique category instead of being ambiguously divided into multiple classes

$$L_e = -\frac{1}{N_T}\sum_{i=1}^{N_T}\sum_{j=1}^{c} z_{ij} \log z_{ij} \quad (13)$$

The total loss for the second stage is

$$L_{RSS} = L_{kl} + \lambda_c L_c + \lambda_e L_e \quad (14)$$

While the entire training is completed, we use the category corresponding to the peak of the distribution $\tilde{y}^c$ as the label, denoted as $Y^n$. Because we cannot guarantee that $Y^n$ is completely reliable, we only keep the samples with $Y^c = Y^n$ and mark the other samples as unlabeled samples.

*3) Instance Training:* The clustering algorithm will cause outliers, and the sample filtering will also generate outlier samples. Most existing methods directly discarded those samples, which may lead to the loss of some difficult samples and only obtain sub-optimal solutions. ACT [20] simply assigned nearest neighbor labels to the corresponding unlabeled sample, which may introduce noise. A more appropriate way is to use them as an individual instance for training.

For the non-outlier samples, the cross-entropy loss $L_{ce}$ is used. It is similar to the form of (1).

For both the non-outlier samples, the batch-hard triplet loss

$$L_{triplet} = \sum_{i=1}^{N_p}\sum_{a=1}^{N_k} [m + \max\|f_i^a - f_i^p\|_2 - \min\|f_i^a - f_i^n\|_2]_+ \quad (15)$$

Especially, for non-outlier samples, $f_i^a, f_i^p, f_i^n$ are features extracted from the anchor, positive and negative samples respectively. $m$ is the margin hyper-parameter. $N_p$ is the number of different identities in a minibatch, $N_k$ is the number of samples in each category. But if $f_i^a$ is a feature of the outlier sample, $f_i^p$ is the feature of the corresponding image transformation sample. The meaning of $f_i^n$ remains unchanged.

The training loss of the entire Main Model is

$$L_{total} = L_{ce} + L_{triplet} \quad (16)$$

## IV. EXPERIMENT

### A. Dataset

We evaluate our method on the three large-scale Re-ID benchmarks: Market-1501 [10], DukeMTMC-ReID [11], [12], CUHK03 [52].

**Market-1501** [10] was collected with 6 different cameras. The training set includes 751 identities and 12,936 images. The test set includes 750 identities, with 3,368 images in the query set and 19,732 images in the gallery set.

**DukeMTMC-ReID** [12] is a sub-dataset of DukeMTMC [11], collected from 8 different cameras. The training set includes 702 identities and 16522 images. The test set includes 702 identities, with 2228 images in the query set and 17611 images in the gallery set.

**CUHK03** [52] was captured by 2 cameras. This dataset was constructed by both manual labeling and DPM. In this work, we experiment on the images detected using DPM. We follow the new train/test evaluation protocol [53] to test our methods. The training set includes 767 identities and 7365 images. The test set includes 700 identities, with 1400 images in the query set and 5332 images in the gallery set.

**Evaluation Protocol**: Cumulative matching characteristic (CMC) and Mean Average Precision (mAP) is used to calculate the performance of Re-ID. All results reported in this paper are under the single-query setting, and no post-processing like re-ranking [53] is applied.

### B. Implementation Details

Firstly, we train the Feature Distribution Alignment module, which can provide more friendly conditions for subsequent clustering process. Then, we alternately train the Main Model and Auxiliary Model, and the Auxiliary Model provides more reliable samples for the Main Model.

For the Feature Distribution Alignment module, we adopt ResNet-50 [54] as the backbone and initialize the model with the parameters pre-trained on ImageNet [55]. We remove the network layer after "Avg pool", and add a fully connected layer with a parameter of 2048, followed by the BN layer, Dropout layer, and Softmax layer. The discriminator is a simple 6-layers perceptron network, extracting features of the "Avg pool" layer for adversarial adaptation. The feature dimension is 2048 in storage and the number of cells is the same as the number of target samples. The updating rate of key memory $\alpha$ is 0.2. The temperature factor $\tau$ is set to 0.05. The neighbor parameter $r_2 = 4$ and $r_1$ is half of $r_2$. The Adam optimizer is used. The learning rate is 0.00035 for fine-tuning layers. Only 10 epochs can achieve great results. The mini-batch size is 64 for both source images and target images.

For the Main Model, which is initialized with the parameters trained on FDA module. We follow the clustering framework in [14] and set the density threshold $p = 1.6 \times 10^{-4}$. The Adam optimizer is used. The learning rate is 0.00035 for overall 40 training epochs.

For the Reliable Sample Selection module, we initialize the Auxiliary Model as the same as the Main Model. For the first stage, we set the initial number of clean samples $k = 12$ and the confidence threshold $\lambda = 0.9$. The first stage lasted 10 epochs. For the second stage, the label softening parameter $\mu = 10$ and the weight parameters $\lambda_c = 0.1, \lambda_e = 0.1$, which lasts for 10 epochs. When the Main Model is trained for 5 epochs, the subnetwork is trained once. When the subnet is trained again, it will be reinitialized.

During the entire training process, the size of input images is $258 \times 128$, using random flipping, random cropping, and random erasure [56] as a means of data augmentation. In testing, we extract the output of the "Avg pool" layer as the image features and adopt the Euclidean distance to measure the similarities between query and gallery images.

### C. Ablation Study

*1) Effectiveness of the Feature Distribution Alignment Module:* As shown in Table I, compared with supervised Re-ID, the effect of unsupervised Re-ID has dropped a lot. For example, when test on Market1501, the Rank-1 accuracy and mAP decrease by 5.2% and 15.1% respectively. The main reason is the domain conflicts. The model cannot extract appropriate features for the target domain samples. To verify the effectiveness of our proposed FDA module. We make the ablation experiments as shown in Table I. It can be observed that equipped with the FDA module ("FDA") can get better results than direct transfer. Specifically, when tested on Market1501 and DukeMTMC the Rank-1 accuracy is improved from 56.4% to 72.1% and 41.1% to 48.0%. As mention above, due to the cross-domain conflicts, it is not reliable to directly utilize the pre-trained source domain model to calculate the sample feature distance in the target domain. The pseudo-labels obtained by clustering will contain a lot of noise, which will seriously affect the recognition performance of the model. It can be seen from Table I that after utilizing the proposed FDA module, the clustering algorithm gets a better condition, the label accuracy is improved, and the recognition effect is also improved. When tested on Market1501, Rank-1 and mAP have reached 90.6% and 74.9%, which is 2.3% and 5.5% higher than the baseline. In the FDA module, we use the adversarial adaptation function and camera-wise contrastive learning. Similar to the use in GAN, the adversarial adaptation function produces a sample feature that is indistinguishable by the domain discriminator to narrow the domain gaps. And through the contrastive learning, we keep the anchor away from other samples. By the camera-wise neighbor mining mechanism, we can search similar samples and narrow their distance. In the end, the id-related samples will be more concentrated and id-unrelated samples will be more scattered.

To further understand the discriminative ability of our FDA module, we utilize t-SNE [57] to visualize the feature distribution by plotting them to a two-dimension map. As shown in Fig. 3, it can be observed that after using the module, the distribution of different classes is more dispersed, and the distribution becomes more even, which is more conducive to the subsequent process.

*2) Effectiveness of the Reliable Sample Selection Module:* Through the clustering algorithm, we obtain noisy labels. Under the guidance of the noisy labels, we can only achieve sub-

TABLE I:
COMPARING THE EFFECTS OF FDA MODULE TESTED ON THE MARKET1501 AND DUKEMTMC DATASETS.

| Methods | Duke ⇒ Market | | Market ⇒ Duke | |
|---|---|---|---|---|
| | R1 | mAP | R1 | mAP |
| Supervised | 93.5 | 84.5 | 84.5 | 71.5 |
| Direct | 56.4 | 28.2 | 41.1 | 24.0 |
| FDA | 72.1 | 44.2 | 48.0 | 26.4 |
| Baseline | 88.3 | 69.4 | 73.8 | 58.8 |
| Baseline + FDA | 90.6 | 74.9 | 78.0 | 61.7 |

"Direct": Train in the source domain and directly test in the target domain. "FDA": Equipped with the Feature Distribution Alignment module. "Baseline": clustering algorithm directly acts on the "Direct". "Baseline + FDA": Clustering algorithm follow the "FDA".

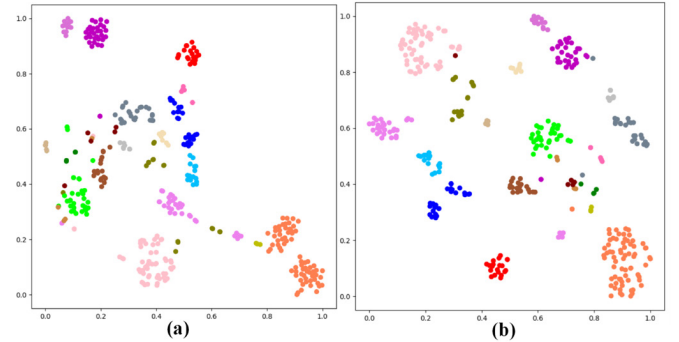

Fig. 3. Visualization of data distributions via t-SNE. The samples are part of the Market1501 training set. The same color represents the images belong to the same category. (a): the model trained only by source domain data. (b): the model equipped with our FDA module.

optimal results. In Table II, we verify the effectiveness of Reliable Sample Selection and instance-level training.

A common assumption is that samples closer to the cluster center are more reliable. Choosing reliable samples to train model according to this criterion can achieve better results. But when the clustering is wrong, the distance criterion alone cannot filter out reliable labels. Because there may be two or more classes in a cluster and they are interlaced with each other. As shown in Table II, the improvement brought by this criterion of filtering reliable samples is very limited. The mAP only increases by 0.4% and 0.6% respectively when tested on the Market1501 and DukeMTMC. If the clustering is wrong, the distance-based sample screening will also be wrong. Because both are based on the distance between samples. As for our method, although the clustering algorithm generates noisy labels, we train the Auxiliary Model parameters and labels in a framework. The accuracy of the label has been improved. The training mechanism of sample iteration prevents the model from being excessively interfered with by noisy labels and the sample filtering mechanism reduces the risk of modifying the original correct label. Because we did not repeatedly use the original inter-sample distance relationship but obtained the relationship between the data structure through a new network. This is very similar to the existing collaborative framework [19], which determines the relationship between samples from two different perspectives. But as the training progresses, the two networks will tend to be similar, at this time the model degenerates into a single network. And our Auxiliary Model

TABLE II:
COMPARING THE EFFECTS RSS MODULE AND THE INSTANCE TRAINING WHEN TESTED ON THE MARKET1501 AND DUKEMTMC DATASETS.

| Methods | Duke ⟹ Market | | Market ⟹ Duke | |
|---|---|---|---|---|
| | R1 | mAP | R1 | mAP |
| Baseline° | 90.6 | 74.9 | 78.0 | 61.7 |
| Baseline° + distance | 90.7 | 75.4 | 78.1 | 62.4 |
| Baseline° + RSS | 90.9 | 77.1 | 78.6 | 63.8 |
| Baseline° + near | 90.0 | 73.5 | 75.6 | 59.4 |
| Baseline° + instance | 90.9 | 75.6 | 77.9 | 62.1 |
| Full model | 91.4 | 77.4 | 79.2 | 64.1 |

"Baseline °": We use "Baseline °" to represent "Baseline + FDA" for convenience. "distance": Select reliable samples based on the distance from the sample to the cluster center. "near": Assigned nearest neighbor labels to the corresponding un-labeled samples. "instance": The un-labeled samples are trained in the instance way.

TABLE III.
EVALUATION WITH DIFFERENT VALUES OF $\tau$ IN EQ. 3

| $\tau$ | Duke ⟹ Market | | Market ⟹ Duke | |
|---|---|---|---|---|
| | R1 | mAP | R1 | mAP |
| 0.01 | 13.6 | 4.8 | 9.0 | 3.2 |
| 0.02 | 23.0 | 8.8 | 16.3 | 6.2 |
| 0.05 | 72.1 | 44.2 | 48.0 | 26.4 |
| 0.1 | 48.9 | 24.9 | 30.2 | 16.9 |
| 0.2 | 33.6 | 15.9 | 18.2 | 8.8 |
| 0.5 | 23.9 | 9.7 | 8.3 | 3.5 |

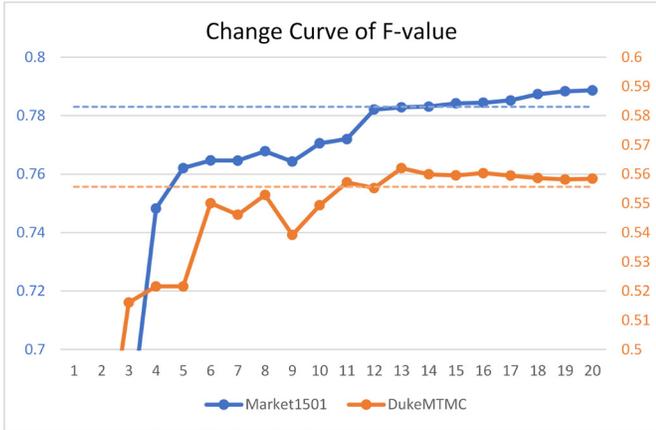

Fig. 4. The F-value of the sample pairs. The solid line represents the change of F-value with training. The dashed line represents the benchmark

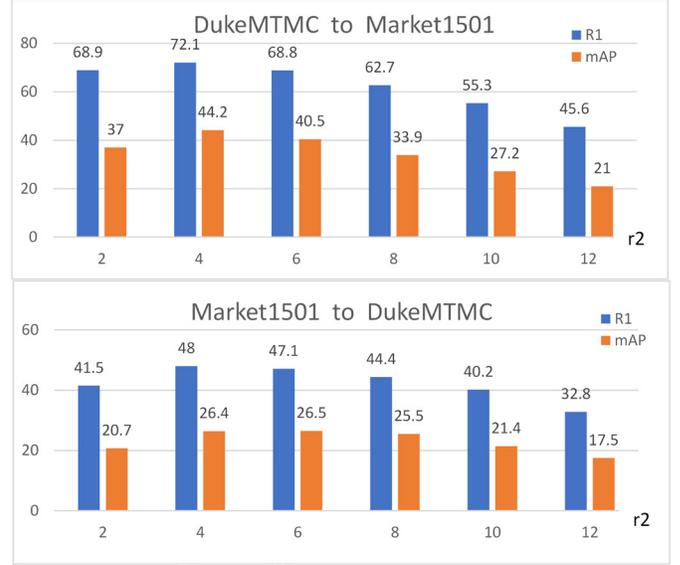

Fig. 5. Evaluation with different values of $r_2$ in Eq. 2.

will be re-initialized every time when it is used, avoiding the problems. Finally, our methods our method has achieved an improvement of 2.2% and 2.1% in mAP when tested on the two datasets.

To intuitively reflect the effectiveness of the RSS module, we draw the F-value in Fig. 4, which is calculated by precision and recall values. Sample pairs belong to the same category in the real labels are regarded as true positive. The dashed line indicates the label accuracy when the module is not used. It can be noticed that in the training process, the F-value gradually increases, which demonstrates that the label noise is indeed decreasing.

The clustering algorithm will produce outliers. In Table II, we compare three methods of handling outliers, namely, direct discard ("baseline°"), nearest neighbor assignment ("baseline° + near") and instance-level training ("baseline° + instance").

We can observe that the nearest neighbor method will reduce the recognition performance. When tested on DukeMTMC, the R1 and mAP reduce 2.4% and 2.3% respectively. Because clustering will produce wrong labels, and simply assigning neighbor labels to outliers will enlarge the error. But the instance-level training method can slightly improve the recognition accuracy. The R1 and mAP have reached 90.9% and 75.6%. This means that outlier samples may be difficult samples and contain information. Direct discarding them will result in a lack of information. Training this part of the sample in instance-level can retain this part of the information without introducing too much noise.

When we combine the three improvement points, we find that the effect has been further improved. When testing on DukeMTMC, the R1 and mAP reached 79.2% and 64.1% respectively. When testing on Market1501, the R1 and mAP obtain 91.4% and 77.4%. It shows that our methods are complementary to a certain extent. The Feature Distribution Alignment module reduces the domain differences and aligns the feature distribution between samples, which creates a friendly condition for the next process. The label correction strategy learns labels from a different perspective, and choose more reliable samples. Although this process will generate more outlier samples, the subsequent instance-level training strategy avoids the side effects of removing samples.

### D. Parameter Analysis

*1) Temperature Factor $\tau$:* In Table III, we investigate the effect of the temperature factor $\tau$ in (3). We noticed that $\tau$ is very sensitive. When its value is too large or too small, the effect will drop drastically. The reason is that a small $\tau$ will make the distribution smooth and approach a uniform distribution, causing all false-positive samples to be close. An overly large $\tau$ will make the distribution sharp and approach a unimodal distribution. The model will degenerate into an instance classification problem, which will inhibit the potential positive sample relationships discovered through camera contrastive learning. The best results are produced when τ is around 0.05.

TABLE IV:
UNSUPERVISED PERSON RE-ID PERFORMANCE COMPARES WITH STATE-OF-THE-ART METHODS ON MARKET1501 AND DUKEMTMC.

| Methods | D $\Rightarrow$ M | | | | M $\Rightarrow$ D | | | |
|---|---|---|---|---|---|---|---|---|
| | R1 | R5 | R10 | mAP | R1 | R5 | R10 | mAP |
| LOMO [26] | 27.2 | 41.6 | 49.1 | 8.0 | 12.3 | 21.3 | 26.6 | 4.8 |
| BOW [10] | 35.8 | 52.4 | 60.3 | 14.8 | 17.1 | 28.8 | 34.9 | 8.3 |
| PTGAN [9] | 38.6 | —— | 66.1 | —— | 27.4 | —— | 50.7 | —— |
| SPGAN+LMP [8] | 58.1 | 76.0 | 82.7 | 26.9 | 46.9 | 62.6 | 68.5 | 26.4 |
| DA-2S [28] | 58.5 | —— | —— | 27.3 | 53.5 | —— | —— | 30.8 |
| CR-GAN + LMP [29] | 64.5 | 79.8 | 85.0 | 33.2 | 56.0 | 70.5 | 74.6 | 33.3 |
| MMFA [6] | 56.7 | 56.7 | 56.7 | 56.7 | 45.3 | 59.8 | 66.33 | 24.7 |
| ECN [40] | 75.1 | 87.6 | 91.6 | 43.0 | 63.3 | 75.8 | 80.4 | 40.4 |
| TJ-AIDL [58] | 58.2 | 74.8 | 81.1 | 26.0 | 44.3 | 59.6 | 65.0 | 23.0 |
| THEROY [14] | 75.8 | 89.5 | 93.2 | 53.7 | 68.4 | 80.1 | 83.5 | 49.0 |
| SSG [18] | 80.0 | 90.0 | 92.4 | 58.3 | 73.0 | 80.6 | 83.2 | 53.4 |
| ATC [20] | 80.5 | —— | —— | 60.6 | 72.4 | —— | —— | 54.5 |
| GDS-H [60] | 81.1 | —— | —— | 61.2 | 73.1 | —— | —— | 55.1 |
| AD-Cluster [59] | 86.7 | 94.4 | 96.5 | 68.3 | 72.6 | 82.5 | 85.5 | 54.1 |
| MMT-500 [19] | 87.7 | 94.9 | 96.9 | 71.2 | 76.8 | 88.0 | 92.2 | 63.1 |
| NRMT [62] | 87.8 | 94.6 | 96.5 | 71.7 | 77.8 | 86.9 | 89.5 | 62.2 |
| Gen [61] | 88.1 | 94.4 | 96.2 | 71.5 | **79.5** | **88.3** | **91.4** | **65.2** |
| DCML [35] | 87.9 | 95.0 | 96.7 | 72.6 | 79.3 | 86.7 | 89.5 | 63.5 |
| **Our** | **91.4** | **96.6** | **97.7** | **77.4** | 79.2 | 87.9 | 91.3 | 64.1 |

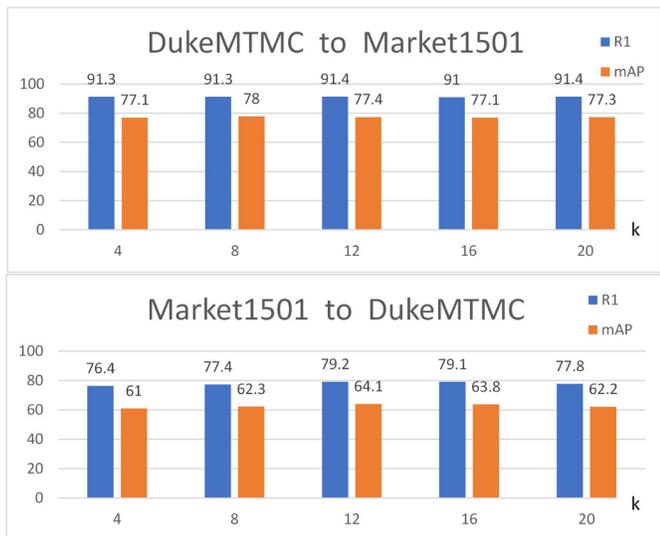

Fig. 6. Evaluation with different values of $k$, which determines the number of samples to initialize the Auxiliary Model.

*2) Neighbor Parameter $r_2$ :* In the Feature Distribution Alignment module, there is an important parameter $r_2$, which decides the number of neighbors. We set $r_1$ equal to half of $r_2$, not making additional discussion. The effect of $r_2$ is shown in Fig. 5. With the increase of $r_2$, the Rank-1 accuracy and mAP first increase and then decrease. The result reaches a peak when the $r_2 = 4$. Through observing the datasets, we find that each pedestrian appears in an average of four cameras in the Market1501 while in three in the DukeMTMC. It is consistent with our experimental results. When $r_2$ is small, the model is not enough to mine all neighbors, and this ability gradually enhance as $r_2$ increases. When $r_2$ is too large, some negative samples will also be included, causing the result to drop.

*3) The Number of Reliable Samples $k$:* In the Auxiliary Model initialization, we need a small number of clean samples to help the model obtain the ability to distinguish instances. In Fig. 6, we investigate the effect of different $k$ values. It can be noticed that $k$ has better robustness. PENCIL selects all sample to initial the Auxiliary Model, but we think which will introduce the noisy labels. The result of $k = 16$ is worse than the result of $k = 12$ to prove this point. But when the value of k is too small, such as $k = 4$, the model may be limited by too few samples to effectively explore the relationship between the data. In the case shown in the paper, $k = 12$ is a good choice.

### E. Comparison with the State-of-the-Art

We conducted 6 sets of experiments on 3 datasets: D $\Rightarrow$ M, M $\Rightarrow$ D, C $\Rightarrow$ M, C $\Rightarrow$ D, D $\Rightarrow$ C, M $\Rightarrow$ C. D, M, C stand for DukeMTMC, Market1501, CUHK03 respectively. We compare with two hand-crafted feature based methods [10], [26], seven reducing domain gap methods [6], [8], [9], [28], [29], [58] and nine clustering-based methods [14], [18-20], [35], [60-62].

TABLE V:
UNSUPERVISED PERSON RE-ID PERFORMANCE COMPARES WITH STATE-OF-THE-ART METHODS WHEN TRANSFER FROM CUHK03 TO MARKET1501 AND DUKEMTMC.

| Methods | C ⇒ D | | C ⇒ M | |
| --- | --- | --- | --- | --- |
| | R1 | mAP | R1 | mAP |
| PTGAN(**O**) [10] | 17.6 | —— | 31.5 | —— |
| DA-2S(**O**) [28] | 47.7 | 27.8 | 57.6 | 28.5 |
| ATC(**O**) [20] | 52.8 | 35.4 | 81.2 | 64.1 |
| GDS-H(**O**) [60] | 64.9 | 45.3 | 84.2 | 66.1 |
| CR-GAN [29] | 46.5 | 26.9 | 58.5 | 30.4 |
| PAST [17] | 69.9 | 51.8 | <u>79.4</u> | 57.3 |
| DCML [35] | <u>72.2</u> | <u>54.6</u> | 78.7 | <u>59.5</u> |
| **Our** | **77.3** | **61.7** | **91.1** | **77.4** |

(**O**) means that the method uses the old evaluation protocol.

TABLE VI:
UNSUPERVISED PERSON RE-ID PERFORMANCE COMPARES WITH STATE-OF-THE-ART METHODS WHEN TRANSFER FROM MARKET1501 AND DUKEMTMC TO CUHK03.

| Methods | D ⇒ C | | M ⇒ C | |
| --- | --- | --- | --- | --- |
| | R1 | mAP | R1 | mAP |
| PTGAN(**O**) [10] | 24.8 | —— | 26.9 | —— |
| THEROY(**O**) [14] | 28.5 | 28.8 | 47.0 | 46.4 |
| ATC(**O**) [20] | 30.6 | 30.0 | 49.5 | 48.9 |
| DA-2S(**O**) [28] | 33.7 | 27.3 | 42.2 | 32.5 |
| GDS-H(**O**) [60] | <u>36.0</u> | <u>34.6</u> | <u>50.2</u> | <u>49.7</u> |
| **Our(O)** | **37.4** | **36.5** | **53.0** | **51.0** |

(**O**) means that the method uses the old evaluation protocol.

*1) Experimental Results on DukeMTMC and Market1501:* In Table IV, we compare our method with state-of-the-art unsupervised learning approaches when tested on Market1501 and DukeMTMC.

First of all, we compare it with the hand-crafted feature based methods, which are trained on neither labeled source data nor the unlabeled target data. It can be observed that among these three categories, the hand-crafted methods perform the worst. Although the hand-crafted features prove to be effective on small datasets, they lack robustness when applied to large and complex datasets.

Secondly, we analyze the domain gap reducing methods. For the image level, the GAN-based approaches tend to generate images with the target domain style and retain the source domain contents. While it is possible to reduce the domain gaps, the performance is affected by the quality of generated images. Crucially, they only learn the traits of the source domain and ignore them in the target domain. Our method outperforms the best GAN method, "CR-GAN + LMP" by 26.9% and 44.2% on Rank-1 and mAP when tested on Market1501. For the feature-level, ECN introduced three underlying invariances to learn the intra-domain variations in the target domain, but the global potential positive samples search is inaccurate. Therefore, it is impossible to effectively mine the relationship between the target domain samples. Some methods [58] also require additional information such as attribute annotations. But our method does not require additional information.

Finally, we focus on clustering-based approaches. The prediction of the pseudo-label and the training of the model alternate. As the training progresses, the label becomes more accurate so that the model can be better trained. At the same time, the model can accurately grasp the data relationship is conducive to extracting more accurate labels. The two complement each other. But the noisy label generated by clustering is a problem to be solved. MMT uses mutual learning and proposes a soft triple loss for training. The two models will gradually converge during the training process and thus cannot guide each other. But our method uses the form of the Main Model and Auxiliary Model. The Auxiliary Model is reinitialized every time, which guarantees the difference from the Main Model. When testing on DukeMTMC, we got 2.4% and 1.0% improvements on R1 and mAP. We noticed that our method obtained better results than Gen when tested on market1501. But when tested on DukeMTMC, mAP is 1.1% worse. The result may be because Gen combined the source domain data and the target domain data for a linear combination, thereby obtaining more data and reducing the risk of overfitting.

In Table V, we also compared the performance of Market1501 and DukeMTMC when the source domain is Cuhk03. The new evaluation protocol [53] is difficult than the original [52] because it uses fewer data to train the model. But we can see from Table V that even in such unfair conditions, our method still far exceeds some methods that use old protocol.

When compared on the new protocol, our method still performs well. For example, when tested on the two datasets, compared to DCML, we achieved 7.1% and 17.9% improvements on mAP respectively. Because DCML selects reliable samples based on the distance from the sample to the corresponding cluster center, and when there is a problem with clustering, this selection method cannot effectively distinguish noisy samples. Our method uses Auxiliary Model to learn and correct labels, screen reliable samples, and effectively avoid the problem of noise accumulation. The instance-level training method also reduces the side effects of removing unreliable samples. In addition, all cluster-based methods have directly used the features extracted by the pre-trained source domain model to calculate the distance matrix of the clustering, the domain differences cause the inaccurate feature distribution. Our method utilizes camera-wise contrastive learning and adversarial adaptation to align the feature distribution of the target domain samples and optimize the distance criterion.

*2）Experimental Results on CUHK03.*

In Table VI, we compare our method with state-of-the-art unsupervised learning approaches when tested on CUHK03.

All methods have used the old protocol, so we keep consistent with them. ACT uses the label of the nearest neighbor as the label of the outlier sample, which will introduce noise. But our method trains the samples in instance level, which does not simply discard this part of samples, nor force them to be classified into a specific category. When testing on Market1501, we got 6.8% and 6.5% improvements on R1 and mAP. GDS-H optimize Re-ID from a global distance-distribution perspective by encouraging the global separation of positive and negative samples, which puts more emphasis on hard mining of samples and has a stronger dependence on clustering accuracy. So noisy labels will have a greater impact on the methods. Our method adopts the Auxiliary Model to learn labels to screen samples, which reduces the influence of noise.

## V. Conclusion

In this paper, we propose an Anti-Noise Learning approach to solve the noise problem in the cross-domain adaptation Re-ID task. Combine the advantage of both the bridging domain gaps and learning target domain feature methods, our method aligns the target domain feature distribution, optimizes the clustering results, and selects reliable samples. Instance-level training of outliers also ensures the effective use of information. Under the guidance of reliable samples, the accuracy of Re-ID is improved, which indicates the importance of denoising. Experimental results validate the effectiveness of the proposed approach in both qualitative and quantitative evaluations.

## REFERENCES

[1] Q. Leng, M. Ye, and Q. Tian, "A Survey of Open-World Person Re-Identification," *IEEE Trans. Circuits Syst. Video Technol.,* vol. 30, no. 4, pp. 1092-1108, Aug. 2020.
[2] Z. Zhu, X. Jiang, F. Zheng, X. Guo, F. Huang, X. Sun, and W.-S. Zheng, "Viewpoint-Aware Loss with Angular Regularization for Person Re-Identification," in *Proc. AAAI,* Feb. 2020, pp. 13114-13121.
[3] W. Li, X. Zhu, and S. Gong, "Harmonious Attention Network for Person Re-identification," in *Proc. Conf. Comput. Vis. Pattern Recognit.,* Jun. 2018, pp. 2285-2294.
[4] J. Lei, L. Niu, H. Fu, B. Peng, Q. Huang, and C. Hou, "Person Re-Identification by Semantic Region Representation and Topology Constraint," *IEEE Trans. Circuits Syst. Video Technol.,* vol. 29, no. 8, pp. 2453-2466, Aug. 2019.
[5] C. Han, R. Zheng, C. Gao, and N. Sang, "Complementation-Reinforced Attention Network for Person Re-Identification," *IEEE Trans. Circuits Syst. Video Technol.,* vol. 30, no. 10, pp. 3433-3445, Oct. 2020.
[6] S. Lin, H. Li, C.-T. Li, and A. C. Kot, "Multi-task Mid-level Feature Alignment Network for Unsupervised Cross-Dataset Person Re-Identification," in *Proc. Brit. Mach. Vis. Conf.,* Sep. 2018, pp.
[7] L. Qi, L. Wang, J. Huo, L. Zhou, Y. Shi, and Y. Gao, "A Novel Unsupervised Camera-Aware Domain Adaptation Framework for Person Re-Identification," in *Proc. Int. Conf. Comput. Vis.,* 27 Oct.-2 Nov. 2019, pp. 8079-8088.
[8] W. Deng, L. Zheng, Q. Ye, G. Kang, Y. Yang, and J. Jiao, "Image-Image Domain Adaptation with Preserved Self-Similarity and Domain-Dissimilarity for Person Re-identification," in *Proc. Conf. Comput. Vis. Pattern Recognit.,* Jun. 2018, pp. 994-1003.
[9] L. Wei, S. Zhang, W. Gao, and Q. Tian, "Person Transfer GAN to Bridge Domain Gap for Person Re-identification," in *Proc. Conf. Comput. Vis. Pattern Recognit.,* Jun. 2018, pp. 79-88.
[10] L. Zheng, L. Shen, L. Tian, S. Wang, J. Wang, and Q. Tian, "Scalable Person Re-identification: A Benchmark," in *Proc. Int. Conf. Comput. Vis.,* Dec. 2015, pp. 1116-1124.
[11] E. Ristani, F. Solera, R. S. Zou, R. Cucchiara, and C. Tomasi, "Performance Measures and a Data Set for Multi-Target, Multi-Camera Tracking," in *Proc. Eur. Conf. Comput. Vis.,* Oct. 2016, pp. 17-35.
[12] Z. Zheng, L. Zheng, and Y. Yang, "Unlabeled Samples Generated by GAN Improve the Person Re-identification Baseline in Vitro," in *Proc. Int. Conf. Comput. Vis.,* Oct. 2017, pp. 3774-3782.
[13] G. Wu, X. Zhu, and S. Gong, "Tracklet Self-Supervised Learning for Unsupervised Person Re-Identification," in *Proc. AAAI,* Feb. 2020, pp.
[14] L. Song, C. Wang, L. Zhang, B. Du, Q. Zhang, C. Huang, and X. Wang, "Unsupervised domain adaptive re-identification: Theory and practice," *Pattern Recognit.,* vol. 102, p. 107173, Jun. 2020.
[15] Y. Lin, X. Dong, L. Zheng, Y. Yan, and Y. Yang, "A Bottom-Up Clustering Approach to Unsupervised Person Re-Identification," in *Proc. AAAI,* Jan. 27 - Feb. 1 2019, pp. 8738-8745.
[16] K. Yi and J. Wu, "Probabilistic End-To-End Noise Correction for Learning With Noisy Labels," in *Proc. Conf. Comput. Vis. Pattern Recognit.,* Jun. 2019, pp. 7010-7018.
[17] X. Zhang, J. Cao, C. Shen, and M. You, "Self-Training With Progressive Augmentation for Unsupervised Cross-Domain Person Re-Identification," in *Proc. Eur. Conf. Comput. Vis.,* 27 Oct.-2 Nov. 2019, pp. 8221-8230.
[18] Y. Fu, Y. Wei, G. Wang, Y. Zhou, H. Shi, U. Uiuc, and T. Huang, "Self-Similarity Grouping: A Simple Unsupervised Cross Domain Adaptation Approach for Person Re-Identification," in *Proc. Int. Conf. Comput. Vis.,* 27 Oct.-2 Nov. 2019, pp. 6111-6120.
[19] Y. Ge, D. Chen, and H. Li, "Mutual Mean-Teaching: Pseudo Label Refinery for Unsupervised Domain Adaptation on Person Re-identification," in *International Conference on Learning Representations* Apr. 2020, pp.
[20] F. Yang, K. Li, Z. Zhong, Z. Luo, X. Sun, H. Cheng, X. Guo, F. Huang, R. Ji, and S. Li, "Asymmetric Co-Teaching for Unsupervised Cross Domain Person Re-Identification," in *Proc. AAAI,* Feb. 2020, pp.
[21] X. Jin, C. Lan, W. Zeng, G. Wei, and Z. Chen, "Semantics-Aligned Representation Learning for Person Re-Identification," in *Proc. AAAI,* Feb. 2020, pp. 11173-11180.
[22] H. Park and B. Ham, "Relation Network for Person Re-Identification," in *Proc. AAAI,* Feb. 2020, pp. 11839-11847.
[23] N. McLaughlin, J. M. d. Rincón, and P. C. Miller, "Video Person Re-Identification for Wide Area Tracking Based on Recurrent Neural Networks," *IEEE Trans. Circuits Syst. Video Technol.,* vol. 29, no. 9, pp. 2613-2626, Aug. 2019.
[24] W. Zhang, S. Hu, K. Liu, and Z.-J. Zha, "Learning Compact Appearance Representation for Video-Based Person Re-Identification," *IEEE Trans. Circuits Syst. Video Technol.,* vol. 29, no. 8, pp. 2442-2452, Aug. 2019.
[25] P. Peng, T. Xiang, Y. Wang, M. Pontil, S. Gong, T. Huang, and Y. Tian, "Unsupervised Cross-Dataset Transfer Learning for Person Re-identification," in *Proc. Conf. Comput. Vis. Pattern Recognit.,* Jun. 2016, pp. 1306-1315.
[26] S. Liao, Y. Hu, Z. Xiangyu, and S. Z. Li, "Person re-identification by Local Maximal Occurrence representation and metric learning," in *Proc. Conf. Comput. Vis. Pattern Recognit.,* Jun. 2015, pp. 2197-2206.
[27] H. Yu, A. Wu, and W. Zheng, "Cross-View Asymmetric Metric Learning for Unsupervised Person Re-Identification," in *Proc. Int. Conf. Comput. Vis.,* Oct. 2017, pp. 994-1002.
[28] Y. Huang, Q. Wu, J. Xu, and Y. Zhong, "SBSGAN: Suppression of Inter-Domain Background Shift for Person Re-Identification," in *Proc. Int. Conf. Comput. Vis.,* 27 Oct.-2 Nov. 2019, pp. 9526-9535.
[29] J. Liu, Z.-J. Zha, D. Chen, R. Hong, and M. Wang, "Adaptive Transfer Network for Cross-Domain Person Re-Identification," in *Proc. Conf. Comput. Vis. Pattern Recognit.,* Jun. 2019, pp. 7202-7211.
[30] Y.-J. Li, F.-E. Yang, Y.-C. Liu, Y.-Y. Yeh, X. Du, and Y.-C. F. Wang, "Adaptation and Re-Identification Network: An Unsupervised Deep Transfer Learning Approach to Person Re-Identification," in *Proc. Conf. Comput. Vis. Pattern Recognit. Workshops.,* Jun. 2018, pp. 172-178.
[31] C.-X. Ren, B.-H. Liang, P. Ge, Y. Zhai, and Z. Lei, "Domain Adaptive Person Re-Identification via Camera Style Generation and Label Propagation," *IEEE Trans. Inf. Forensics Secur.,* vol. 15, pp. 1290-1302, / 2020.
[32] L. Qi, L. Wang, J. Huo, Y. Shi, and Y. Gao, "Progressive Cross-Camera Soft-Label Learning for Semi-Supervised Person Re-Identification," *IEEE Trans. Circuits Syst. Video Technol.,* vol. 30, no. 9, pp. 2815-2829, Sep. 2020.
[33] D. Kumar, P. Siva, P. Marchwica, and A. Wong, "Unsupervised Domain Adaptation in Person re-ID via k-Reciprocal Clustering and Large-Scale Heterogeneous Environment Synthesis," in *IEEE Winter Conference on Applications of Computer Vision* Mar. 2020, pp. 2634-2643.
[34] Y. Wu, Y. Lin, X. Dong, Y. Yan, W. Bian, and Y. Yang, "Progressive Learning for Person Re-Identification With One Example," *IEEE Trans. Image Process.,* vol. 28, no. 6, pp. 2872-2881, Sep. 2019.
[35] G. Chen, Y. Lu, J. Lu, and J. Zhou, "Deep Credible Metric Learning for Unsupervised Domain Adaptation Person Re-identification," in *Proc. Eur. Conf. Comput. Vis.,* Aug. 2020, pp. 643-659.

[36] B. Han, I. W. Tsang, L. Chen, C. P. Yu, and S.-F. Fung, "Progressive Stochastic Learning for Noisy Labels," *IEEE Trans. Neural Networks Learn. Syst.,* vol. 29, no. 10, pp. 5136-5148, Mar. 2018.

[37] H.-S. Chang, E. G. Learned-Miller, and A. McCallum, "Active Bias: Training More Accurate Neural Networks by Emphasizing High Variance Samples," in *Proc. Conf. Neural Inf. Process.,* Dec. 2017, pp. 1002-1012.

[38] K. Sohn, "Improved Deep Metric Learning with Multi-class N-pair Loss Objective," in *Proc. Conf. Neural Inf. Process.,* Dec. 2016, pp. 1849-1857.

[39] R. D. Hjelm, A. Fedorov, S. Lavoie-Marchildon, K. Grewal, P. Bachman, A. Trischler, and Y. Bengio, "Learning deep representations by mutual information estimation and maximization," in *International Conference on Learning Representations* May. 2019, pp.

[40] Z. Zhong, L. Zheng, Z. Luo, S. Li, and Y. Yang, "Invariance Matters: Exemplar Memory for Domain Adaptive Person Re-Identification," in *Proc. Int. Conf. Comput. Vis.,* Jun. 2019, pp. 598-607.

[41] J. Goldberger and E. Ben-Reuven, "Training deep neural-networks using a noise adaptation layer " in *International Conference on Learning Representations* Apr. 2017, pp.

[42] A. P. Dawid and A. M. Skene, "Maximum Likelihood Estimation of Observer Error-Rates Using the EM Algorithm," *Journal of the Royal Statistical Society: Series C (Applied Statistics),* vol. 28, no. 1, 1979.

[43] J. Lee, D. Yoo, and H.-E. Kim, "Photometric Transformer Networks and Label Adjustment for Breast Density Prediction," in *Proc. Conf. Comput. Vis. Pattern Recognit. Workshops.,* 27 Oct.-2 Nov. 2019, pp. 460-466.

[44] L. Jiang, Z. Zhou, T. Leung, L.-J. Li, and F.-F. Li, "MentorNet: Learning Data-Driven Curriculum for Very Deep Neural Networks on Corrupted Labels," in *Proc. Int. Conf. Mach. Learn.,* Jul. 2018, pp.

[45] A. Ghosh, H. Kumar, and P. S. Sastry, "Robust Loss Functions under Label Noise for Deep Neural Networks," in *Proc. AAAI,* Feb. 2017, pp. 1919-1925.

[46] Z. Zhang and M. Sabuncu, "Generalized Cross Entropy Loss for Training Deep Neural Networks with Noisy Labels," in *Proc. Conf. Neural Inf. Process.,* Dec. 2018, pp.

[47] Z. Wu, Y. Xiong, S. X. Yu, and D. Lin, "Unsupervised Feature Learning via Non-parametric Instance Discrimination," in *Proc. Conf. Comput. Vis. Pattern Recognit.,* Jun. 2018, pp. 3733-3742.

[48] A. Wu, W. Zheng, and J. Lai, "Unsupervised Person Re-Identification by Camera-Aware Similarity Consistency Learning," in *Proc. Int. Conf. Comput. Vis.,* 27 Oct.-2 Nov. 2019, pp. 6921-6930.

[49] M. Ester, H.-P. Kriegel, J. Sander, and X. Xu, "A density-based algorithm for discovering clusters a density-based algorithm for discovering clusters in large spatial databases with noise," in *Proc. of the Second Int. Conf. on Knowledge Discovery and Data Mining* 1996, pp. 226-231.

[50] B. Han, Q. Yao, X. Yu, G. Niu, M. Xu, W. Hu, I. Tsang, and M. Sugiyama, "Co-teaching: Robust Training of Deep Neural Networks with Extremely Noisy Labels," in *Proc. Conf. Neural Inf. Process.,* Dec. 2018, pp. 8536-8546.

[51] A. Hermans, L. Beyer, and B. Leibe, "In Defense of the Triplet Loss for Person Re-Identification," *CoRR,* vol. abs/1703.07737, 2017.

[52] W. Li, R. Zhao, T. Xiao, and X. Wang, "DeepReID: Deep Filter Pairing Neural Network for Person Re-identification," in *Proc. Conf. Comput. Vis. Pattern Recognit.,* Jun. 2014, pp. 152-159.

[53] Z. Zhong, L. Zheng, D. Cao, and S. Li, "Re-ranking Person Re-identification with k-Reciprocal Encoding," in *Proc. Conf. Comput. Vis. Pattern Recognit.,* Jul. 2017, pp. 3652-3661.

[54] K. He, X. Zhang, S. Ren, and J. Sun, "Deep Residual Learning for Image Recognition," in *Proc. Conf. Comput. Vis. Pattern Recognit.,* Jun. 2016, pp. 770-778.

[55] J. Deng, W. Dong, R. Socher, L. Li, L. Kai, and F.-F. Li, "ImageNet: A large-scale hierarchical image database," in *Proc. Conf. Comput. Vis. Pattern Recognit.,* Jun. 2009, pp. 248-255.

[56] Z. Zhong, L. Zheng, G. Kang, S. Li, and Y. Yang, "Random Erasing Data Augmentation," in *Proc. AAAI,* Feb. 2020, pp. 13001-13008.

[57] v. d. M. Laurens and H. Geoffrey, "Visualizing Data using t-SNE," *Journal of Machine Learning Research,* vol. 9, no. 2605, pp. 2579-2605, 2008.

[58] J. Wang, X. Zhu, S. Gong, and W. Li, "Transferable Joint Attribute-Identity Deep Learning for Unsupervised Person Re-identification," in *Proc. Conf. Comput. Vis. Pattern Recognit.,* Jun. 2018, pp. 2275-2284.

[59] Y. Zhai, S. Lu, Q. Ye, X. Shan, J. Chen, R. Ji, and Y. Tian, "AD-Cluster: Augmented Discriminative Clustering for Domain Adaptive Person Re-Identification," in *Proc. Conf. Comput. Vis. Pattern Recognit.,* Jun. 2020, pp. 9018-9027.

[60] X. Jin, C. Lan, W. Zeng, and Z. Chen, "Global Distance-Distributions Separation for Unsupervised Person Re-identification," in *Proc. Eur. Conf. Comput. Vis.,* Aug. 2020, pp. 735-751.

[61] C. Luo, C. Song, and Z. Zhang, "Generalizing Person Re-Identification by Camera-Aware Invariance Learning and Cross-Domain Mixup," in *Proc. Eur. Conf. Comput. Vis.,* Aug. 2020, pp. 224-241.

[62] F. Zhao, S. Liao, G.-S. Xie, J. Zhao, K. Zhang, and L. Shao, "Unsupervised Domain Adaptation with Noise Resistible Mutual-Training for Person Re-identification," in *Proc. Eur. Conf. Comput. Vis.,* Aug. 2020, pp. 526-544.